\documentclass[lettersize,journal]{IEEEtran}
\UseRawInputEncoding
\IEEEoverridecommandlockouts
\usepackage[letterpaper,top=54pt,bottom=54pt,left=54pt,right=54pt]{geometry}

\usepackage{times}
\usepackage{amsmath,amssymb,amsopn,amstext,amsfonts}
\usepackage{cancel}
\usepackage[space]{cite}
\usepackage{balance}
\usepackage{color}
\usepackage{mathtools}
\usepackage[ruled,vlined,linesnumbered]{algorithm2e}

\usepackage{bm}

\usepackage{diagbox}
\usepackage{float}
\usepackage{pifont}
\usepackage{amsmath}
\usepackage{multirow}
\usepackage{booktabs}
\usepackage{url}
\usepackage{svg}
\usepackage{threeparttable}
\usepackage[linkcolor=black,citecolor=black,urlcolor=black,colorlinks=true]{hyperref}
\usepackage{makecell}

\usepackage{array}
\usepackage{marvosym}

\DeclareMathOperator*{\argmin}{arg\,min}

\DeclareGraphicsExtensions{.png,.jpg,.eps,.pdf}
\DeclareGraphicsExtensions{.pdf}

\title{\LARGE \bf
Number Adaptive Formation Flight Planning via Affine Deformable Guidance in Narrow Environments
}

\author{Yuan Zhou, Jialiang Hou, Guangtong Xu, and Fei Gao% <-this % stops a space
\thanks{This work was supported by the National Key R\&D Program of China under Grant No. 2023YFB4706600, the Zhejiang Provincial Science and Technology Plan Project under Grant No. 2024C01170 and the National Natural Science Foundation of China under Grant Nos. 62322314 \& 62203256. \emph{(Corresponding Author: Jialiang Hou; Guangtong Xu; Fei Gao.)}}
\thanks{Yuan Zhou, Jialiang Hou, and Fei Gao are with the Institute of Cyber-Systems and Control, College of Control Science and Engineering, Zhejiang University, Hangzhou 310027, China, and also with the Huzhou Institute, Zhejiang University, Huzhou 313000, China. (e-mail: \{y2zhou, fgaoaa\}@zju.edu.cn, jlhou8@gmail.com). Guangtong Xu is with the Huzhou Institute, Zhejiang University, Huzhou 313000, China. (e-mail: guangtong\_xu@163.com)}
}

\begin{document}

\maketitle

\begin{abstract}
Formation maintenance with varying number of drones in narrow environments hinders the convergence of planning to the desired configurations. To address this challenge, this paper proposes a formation planning method guided by Deformable Virtual Structures (DVS) with continuous spatiotemporal transformation. Firstly, to satisfy swarm safety distance and preserve formation shape filling integrity for irregular formation geometries, we employ Lloyd algorithm for uniform $\underline{PA}$rtitioning and Hungarian algorithm for $\underline{AS}$signment (PAAS) in DVS. Subsequently, a spatiotemporal trajectory involving DVS is planned using primitive-based path search and nonlinear trajectory optimization. The DVS trajectory achieves adaptive transitions with respect to a varying number of drones while ensuring adaptability to narrow environments through affine transformation. Finally, each agent conducts distributed trajectory planning guided by desired spatiotemporal positions within the DVS, while incorporating collision avoidance and dynamic feasibility requirements. Our method enables  up to 15\% of swarm numbers to join or leave in cluttered environments while rapidly restoring the desired formation shape in simulation. Compared to cutting-edge formation planning method, we demonstrate rapid formation recovery capacity and environmental adaptability. Real-world experiments validate the effectiveness and resilience of our formation planning method.
\end{abstract}

\section{Introduction}
\label{sec:introduction}
\begin{figure}[!t]
    \centering
    \includegraphics[width=3.4in]{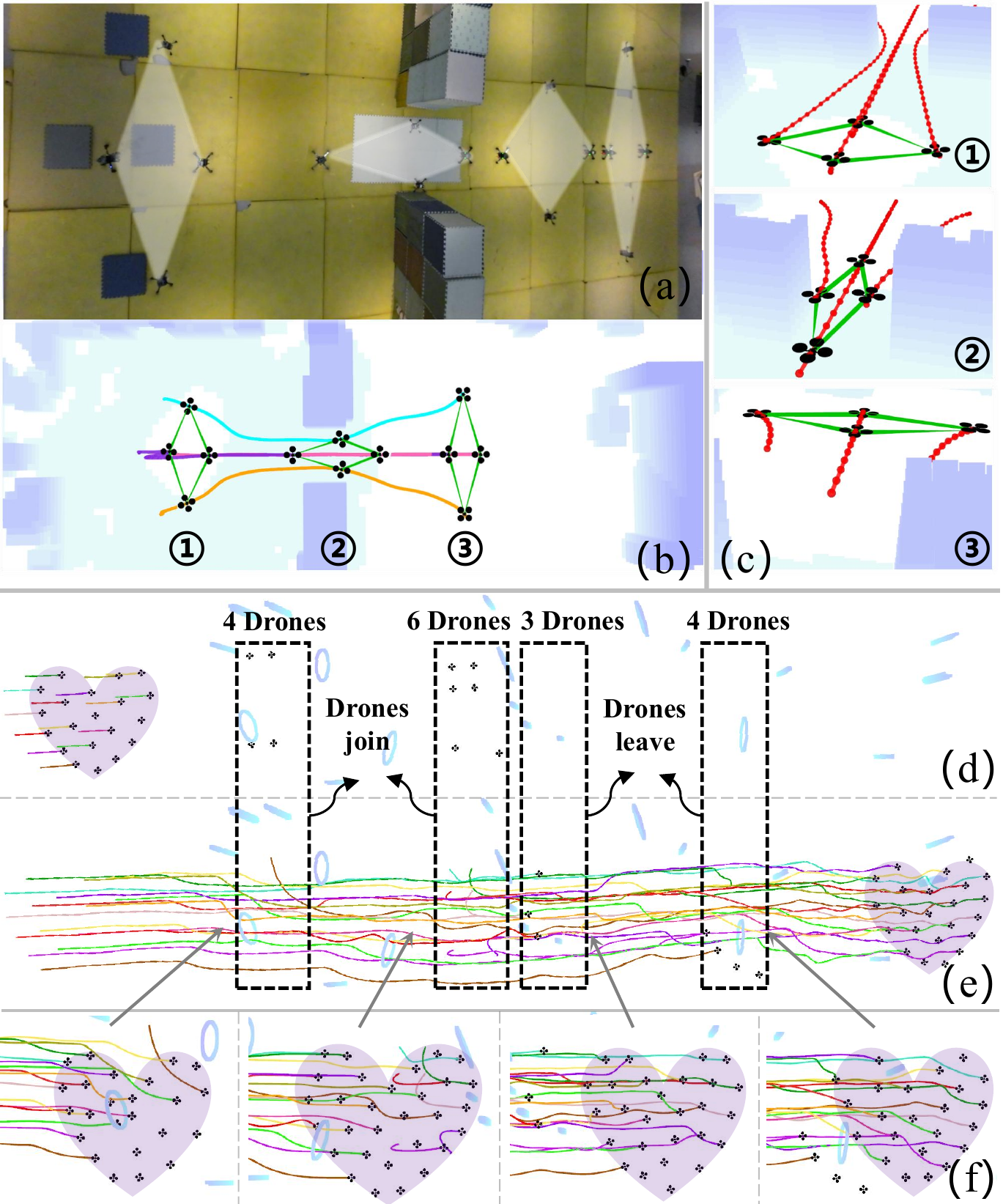}
    \vspace{-0.6cm}
    \caption{
    Formation flight planning via deformation guidance. (a)\&(b) Real world experiment and Rviz diagrams that a formation with 4 drones is traversing a narrow gap. (c) Screenshots and Rviz diagrams of corresponding formations at different time. (d) Initial formation of 20 agents with 4 agents and 6 agents awaiting join in the formation, respectively. (e) Formation flight to the destination. (f) Snapshot of the initial formation, Snapshot of 4 agents joining the formation, Snapshot of 6 agents joining the formation, Snapshot of 3 agents leaving the formation, Snapshot of 4 agents leaving the formation, respectively.
    }
    \vspace{-0.7cm}
    \label{fig1}
\end{figure}

In recent years, formation flight becomes the foundation requirement for aerial swarms in practical applications, such as collaborative exploration \cite{exploration}, light show \cite{lightshow}, search and rescue \cite{search2}. For large-scale swarms \cite{1000agents}, \cite{wu}, formation inevitably encounters agent loss in narrow environments \cite{resilient1, resilient2, zhaoswarm}. Naturally, the integration of new members can restore the impaired formation and enhance the efficiency of mission execution \cite{adaptive}, \cite{merging}. Therefore, dynamic adaptation of formation reconfiguration for changes in the number of drones is essential to ensure operational resilience under unforeseen disturbance. Besides, formation needs to experience adaptively overall shape transformations to navigate through narrow environments and maintain its configuration as much as possible to keep resilience \cite{c1, c16, zhao2018affine, jiaozi}.

When the number of drones in formations varies, the abrupt variation of formation cooperative constraints tends to degrade the feasibility of trajectory optimization. In narrow environments, the inherent conflict between formation maintenance and collision avoidance also deteriorates the feasibility, resulting in no solution for the classical planning method without global formation adjustment mechanism.
% When swarm size variations occur, it is necessary to partition and assign desired position for agents, while ensuring formation shape rescales dynamically to maintain swarm collision avoidance and coverage integrity. Furthermore, rapid formation recovery is essential to guarantee cooperative performance during formation reconfiguration. 

Numerous works have demonstrated formation systems with swarm number variations.
Artificial Potential Fields (APF) \cite{1000agents}, \cite{zhaoswarm}, \cite{shape1} are commonly used to guide the formation generation with variable swarm numbers. This approach eliminates the need for explicit assignment of desired positions, enabling adaptive number variation of formation. However, the slow convergence rate and inadequate swarm collision avoidance hinder its applicability in real-time formation planning. Although explicit position and assignment during number variations enable rapid formation recovery \cite{c17}, \cite{mora_form}, \cite{cat}, they often fail to guarantee robust performance in narrow environments, where collision avoidance remains challenging during formation recovery. 
Adaptive deformation of the formation shape is widely adopted for safe and rapid navigation through obstacle environments \cite{c1}, \cite{zhao2018affine}. Nevertheless, distributed cooperative planning is insufficient without effective guidance for formation in narrow environments. Consequently, adjusting the overall deformation of the swarm provides an effective resolution \cite{c1}. However, this category of approaches is limited by either locally reactive mechanisms \cite{zhao2018affine} and simply adjusts the local target that hinders formation guidance and environmental adaptability \cite{c1}, or an excessive pursuit of collision-free passage that severely compromises formation maintenance integrity while neglecting post-maneuver recovery capability. These methods often struggle to rapidly restore desired formations in cluttered environments. Furthermore, most studies of formation planning do not address adaptive adjustment with variations in swarm number \cite{c1}, \cite{c9}, \cite{consence}.

To achieve real-time formation flight with a variable number of drones, we propose a number adaptive formation planning system. According to the desired formation shape, we employ explicit partitioning and assignment to designate formation positions for the swarm. We focus on designing a Deformable Virtual Structures (DVS) trajectory with spatiotemporal motion and deformation that can effectively guide formation recovery and maintenance. Specifically, we introduce virtual rigid structures with two additional degrees of freedom, including scaling and affine transformation in DVS. The concept of virtual structures has been applied in multiple fields, but predominantly uses rigid virtual structures which are unable to adaptively deform \cite{vrb, vrb1, vrb2}. In this work, we design a spatiotemporal trajectory optimization method for DVS, which utilizes compact piecewise polynomial representations in Cartesian and deformable spaces to formulate nonlinear optimization problems. This enables adaptive adjustment of the scale of formations according to swarm number variations, while permitting appropriate affine deformation of the virtual structures to facilitate the formations to navigate in narrow environments. Integrating PAAS and DVS with spatiotemporal transformation, we designate this framework as Deformation Guidance (DG), and agents perform distributed trajectory planning under the guidance of DG. Extensive simulation and real-world experiments demonstrate that our method supports variable number formation planning while achieving a high success rate in navigating through narrow environments.

The contributions of this paper are as follows:

\begin{enumerate}
% \item We design a spatiotemporal trajectory optimization method for DVS. We employ a primitive DVS path search in the front-end to initialize the DVS trajectory generation, which utilizes compact piecewise polynomial representations in Cartesian and deformable spaces to formulate nonlinear optimization problems. By adding constraints that satisfy the constraint of agent dynamics within the DVS, we guarantee robust convergence to an optimal trajectory for guiding formation coordination.
\item To satisfy swarm safety distance and preserve shape filling integrity for irregular formation geometries, we implement the Lloyd algorithm for uniform partitioning in formation planning with variable number, with assignment via the Hungarian algorithm. We experimentally verified that the computational efficiency of PAAS is sufficient to achieve real-time planning.

\item We propose a spatiotemporal trajectory optimization method for DVS with scaling and affine transformation to guide formation planning in narrow environments. We employ a primitive path search to efficiently obtain high-dimensional DVS paths as initial values for DVS trajectory optimization, which can be transformed into unconstrained optimization problems for real-time solution.

\item We incorporate DVS-based spatiotemporal formation constraint in distributed trajectory optimization to achieve rapid formation recovery and enhanced environmental adaptability. Extensive experiments and benchmarks validate the efficiency and advantages of our system.
\end{enumerate}

\section{Related Work}
\label{sec:related_work}
\begin{figure*}[!t]
    \centering
    \vspace{-0.6cm}
    \includegraphics[width=7.0in]{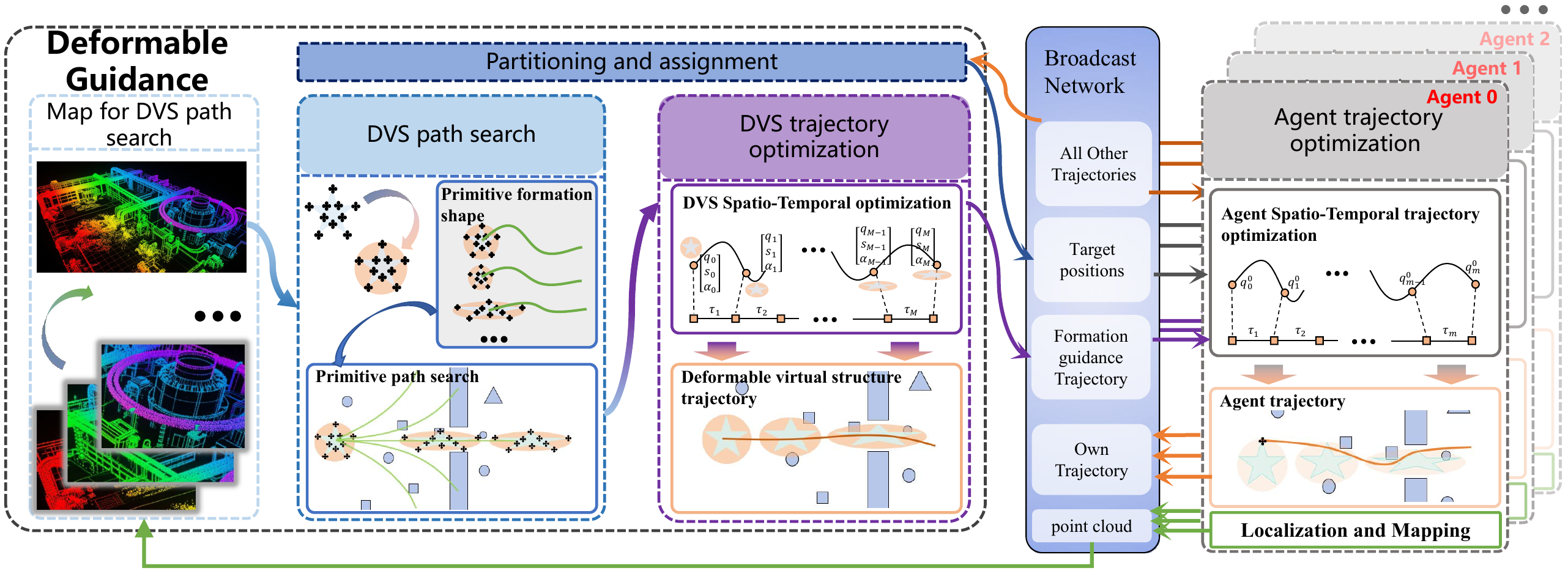}
    \vspace{-0.6cm}
    \caption{
    Our system framework. The system comprises the Deformable Guidance (DG), agent trajectory optimization, and wireless communication. The DG uses its own perception point cloud and point clouds broadcasted by drones to construct an occupancy grid map, which are used for path search of DVS. The DG integrates the formation position partitioning and assignment, path search for the DVS, and the spatiotemporal trajectory optimization of the DVS. Agents achieve distributed trajectory planning guided by DG and broadcast point cloud to DG.
    }
    \vspace{-0.3cm}
    \label{framework}
    \vspace{-0.2cm}
\end{figure*}

Numerous studies focus on formation planning in obstacle environments \cite{c1}, \cite{c16}, \cite{zhao2018affine}. We review existing research on collision avoidance and formation maintenance using adaptive deformation. In addition, a survey of approaches for handling swarm number variations is presented.
\subsection{Formation Planning with Adaptive Deformation}
Numerous studies focus on achieving formation navigation, including virtual structures \cite{vrb}, leader-follower \cite{leaader}, and consensus-based local control laws \cite{consence}. However, most disregard obstacle environments and are prone to deadlocks in large-scale swarms, failing to meet the implicit requirement of scalable formation coordination under swarm size variations. 
Several works explore formation shape deformation to improve environmental adaptability. Mora et al. \cite{c16} employ a convex polytope to represent the formation, deforming the convex polytope within safe convex corridors to ensure collision avoidance. However, they focus on obstacle avoidance, which leads to unrestricted deformation. To balance collision avoidance and formation maintenance, Zhao et al. \cite{zhao2018affine} adopt affine transformations for collision avoidance but rely on predefined global trajectories, limiting adaptability in complex environments. Similarly, Peng et al. \cite{peng} utilize affine deformation but need to centrally optimize trajectories for all agents, making it difficult to meet the potential demand for large-scale formation planning. Quan et al. \cite{c1} propose a graph-based metric to quantify the similarity between formations and integrate it into the local trajectory optimization, using single-readjustment of local formation endpoints to guide the overall formation through narrow gaps. However, they ignore swarm number variations, and the limited guiding ability of the single-step strategy leads to a low success rate in collision-free formation planning.

\subsection{Number Adaptive Formation System}
Number adaptation is a fundamental requirement in the formation system. Markov-based probability distribution methods divide the desired shape into probabilistic boxes \cite{1mswarm}, \cite{markov}, controlling the swarm to converge to the target distribution. Although these methods theoretically have adaptability to number variations, they exhibit slow formation recovery and are primarily limited to obstacle-free environments, making them unsuitable for complex environments. Another category of methods employs artificial potential fields to form formation \cite{1000agents}, \cite{zhaoswarm}, \cite{shape1}. These methods rely on continuous potential fields, inherently permitting swarm number variations without requiring explicit formation position partitioning and assignment. Zhao et al. \cite{zhaoswarm} employ APF to encode desired formations, combining shape-attractive and swarm repulsive forces for uniform distribution. However, these methods demonstrate slow convergence to target shapes. To achieve rapid formation generation, most existing methods employ algorithms to explicitly handle formation position partitioning and assignment tasks \cite{c17}, \cite{mora_form}, \cite{vis_form}. Morgan et al. \cite{c17} determine the number of agents remaining and assign the target based on bidding information, which can adaptively handle swarm number variations. Mora et al. \cite{mora_form} utilize Voronoi partitioning and Hungarian algorithm, decoupling the relative position computation from assignment. Although being capable of rapidly generating visually optimal formations based on the current swarm number, they omit full agent dynamics and disregard obstacles.

\section{SYSTEM OVERVIEW}
The proposed system framework is illustrated in Fig. \ref{framework}. The left of the framework delineates the DG algorithm pipeline, the right side represents distributed single agent trajectory planning, and the information sharing among agents is transmitted through wireless networks. 
\subsection{Deformable Guidance Generation}
Given that all agents within the swarm possess identical system configurations, any agent can act as the role of DG. Initially, DG uses its own perception point clouds and point clouds broadcasted by some drones to construct an occupancy grid map. DG collects the trajectories of all agents, extract their current positions, and accordingly performs formation position partitioning and assignment based on the desired formation shape by Lloyd and Hungarian algorithms (Sect. \ref{assig}). Subsequently, DG performs the path search of DVS on the generated grid map to obtain the initial DVS path (Sect. \ref{frond_end}). This preliminary path serves as the initial condition for the trajectory optimization of DVS, which yields a continuously deformable DVS trajectory by solving a nonlinear optimization problem (Sect. \ref{mid_end}). 
\subsection{Agent distributed Trajectory Optimization}
The DVS trajectory and the outcome of PAAS are broadcast to all other agents. They serve as formation constraints during agent trajectory optimization to generate formation-coordinated optimal trajectories (Sect. \ref{back_end}). We adopt MINCO \cite{c32} to parameterize agent trajectories, where each agent distributively optimizes its trajectory by holistically considering dynamic constraints, collision avoidance, and formation coordination.
\subsection{State Estimation and Mapping}
To enable autonomous perception and mapping, we deploy an omnidirectional LiDAR and onboard IMU as real-time sensing modules. Robust localization is achieved via FAST-LIO2 \cite{fastlio}， which is an open-source LiDAR-inertial odometry (LIO) framework. For single agent obstacle avoidance, we construct an Euclidean Signed Distance Field (ESDF) map \cite{esdf} using point clouds from the LiDAR perception for agent trajectory optimization. Then, we selectively transmit point clouds from agents on the formation outline. These agents transmit only the point clouds perceived within their 120 degree FoV to DG for building sufficient occupancy grid maps, which is used to search the DVS path.
\subsection{Swarm wireless Communication}
Within our framework, the transmission of agent trajectories, point clouds, and PAAS results necessitates highly reliable communication conditions. We utilize 5G networks as the communication backbone for the system. Within our framework, wireless transmission of point clouds dominates bandwidth consumption. To demonstrate that network conditions suffice for point cloud communication requirements, we conducted practical tests in the experimental section, documenting transmission latency under varying number of points (Sect. \ref{point_exp}). Based on the experimental results, we select a broadcast frequency of 10 Hz for point clouds with 500 points per broadcast.

\section{NUMBER AFFINE ADAPTIVE DEFORMABLE GUIDANCE}
\subsection{Formation Partitioning and Assignment}
\label{assig}
\subsubsection{Formation Partitioning}
When the swarm number changes, the desired formation pattern needs to be preserved by dynamically reassigning relative positions based on the original shape. Due to the presence of downwash flow, for a desired formation shape, we adopt a uniform slicing method along the z-axis to generate $k$ planes, where the slice interval equals the safety distance required for downwash flow mitigation. For each sliced plane, we characterize it using a polygon representation. The 
ratio of agent number in different layers $n_1:n_2:\dots:n_k$ is close to the ratio of each layer area $a(W_1):a(W_2):\dots:a(W_k)$. Then, the number of agents in different layers is allocated as $n_1:n_2:\dots:n_k$, where $\sum_{m=0}^{k} n_m=n$, $n$ is the current agent number. To ensure that the new formation closely approximates the original geometry, we reformulate this as a coverage optimization problem \cite{mora_form}. Given a target shape of the $m^{th}$ layer, the objective is to determine the optimal positions for $n_m$ agents through constrained optimization:

\begin{align}
		\min_{\mathbf{p_{c}}} J &= 
		\int_{W_m} \|\mathbf{q}_m-\mathbf{p}_{c} \|\rho(\mathbf{q}_m)d\mathbf{q}_m, \nonumber \\
         &p_{i,c} \in \mathbf{p}_{c}, \forall i \in \{1,...,{n_m}\}.
\end{align}
The set ${W_m}$ is the region of the $m^{th}$ layer shape. $\mathbf{p}_{c}$ is the desired formation position. $p_{i,c}$ is the expected relative position of agent $i$ in the centroid of the DVS before deformation. $\rho(\mathbf{q}_m)$ is the a mass density function that takes positive values inside ${W_m}$ and zero outside of ${W_m}$.
To obtain the desired positions of the target points, this problem can be solved using the Lloyd algorithm. To accelerate convergence, we initialize the algorithm by partitioning the target shape with a rectangular grid to generate initial seed points. In each iteration step, the optimal spatial partitioning for the current samples is computed \cite{mora_form}. We employ a centralized Lloyd algorithm with a computational complexity of $\mathcal O(n \log n)$. 
\subsubsection{Formation Assignment}
\begin{figure}[!t]
    \centering
    \includegraphics[width=3.2in]{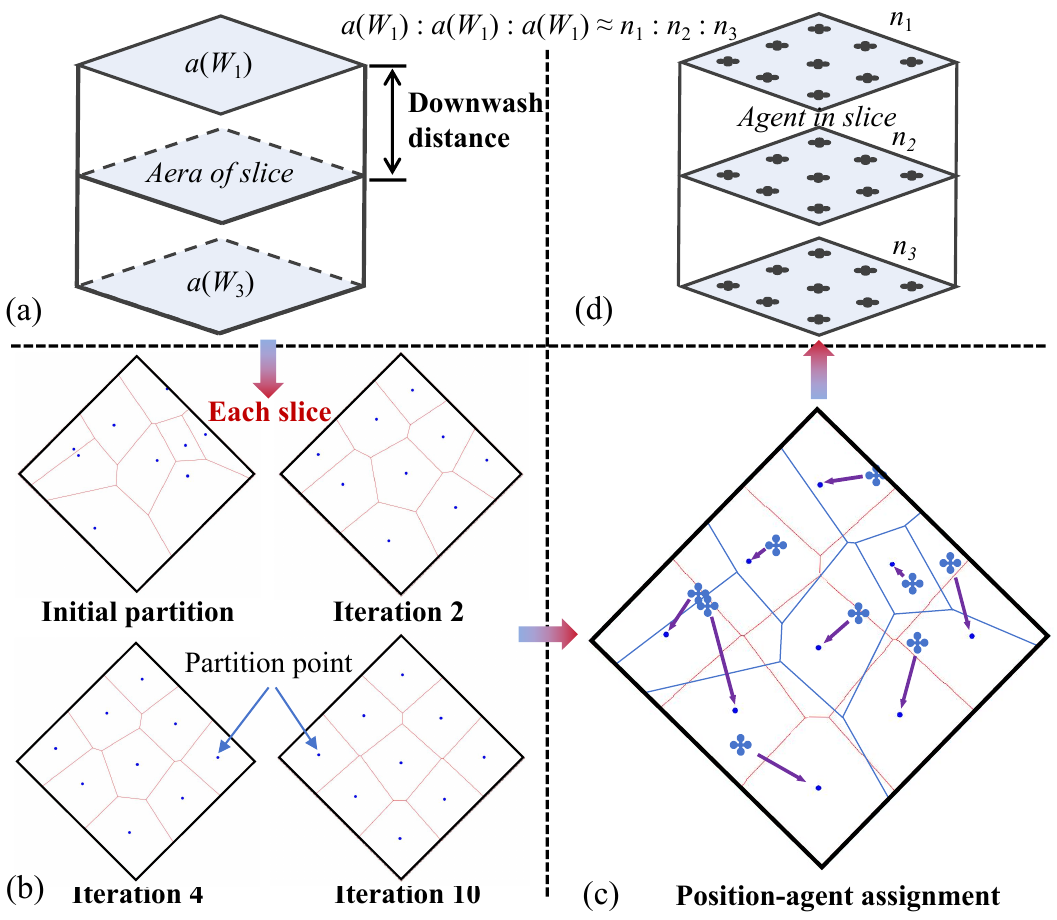}
    \vspace{-0.4cm}
    \caption{
     Formation shape partitioning and assignment. (a) The desired formation shape is sliced into equally spaced layers based on downwash safety distance. (b) It demonstrates the partitioning results and distribution of partition centroids at the initial stage, second iteration, fourth iteration, and final iteration during the partitioning process of a layer. (c) The Hungarian algorithm is employed to assign current position of agents to the partitioned centroids. (d) The partitioned and assigned target formation.
    }
    \vspace{-0.3cm}
    \label{alas}
    \vspace{-0.2cm}
\end{figure}

Finally, after obtaining the target positions $p_{i,c}$, we employ the Hungarian algorithm to further optimize drone position assignments
\begin{align}
v =  \argmin_{v} \sum_{i=1}^{n}\|p_{i,c}- p_i\|^2,
\end{align}
where $v$ is the assignment function and $p_i$ is the position of the agent $i$. We use the centralized Hungarian algorithm and the computational complexity is $\mathcal O(n^3)$. The PAAS is demonstrated through an example presented in Fig. \ref{alas}.
% thereby minimizing collision risks from inter-agent trajectory crossings and reducing energy consumption.
\subsection{Path Search of Deformable Guidance}
\label{frond_end}
\begin{figure}[!t]
    \centering
    \includegraphics[width=3.5in]{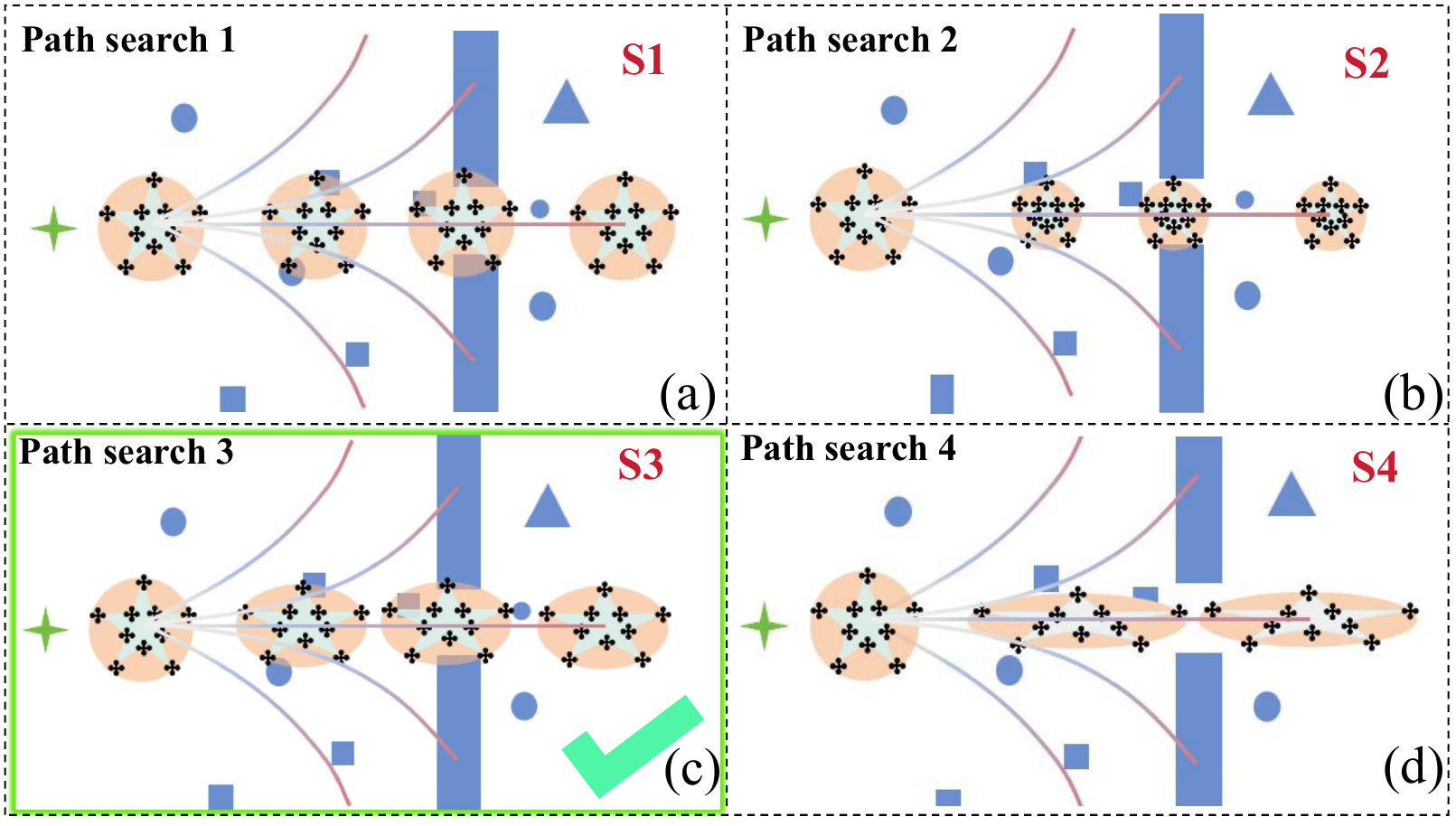}
    \vspace{-0.6cm}
    \caption{
    Schematic of primitive path selection in narrow environments. (a)-(d) At a given timestep, four representative examples demonstrate the selection process and scoring of DG's primitive topologies in an obstacle environment. Path search 1 exhibits excessive $c_o$, path search 2 exhibits excessive $c_r$, path search 4 suffers from overlarge $c_\alpha$, and topologies in other directions either exhibit excessive $c_o$ or overlarge $c_e$. As $S_3>Max\{ S_1, S_2, S_4\}$, path search 3 achieves the highest score and is selected.
    }
    \vspace{-0.4cm}
    \label{frond}
\end{figure}

To facilitate obstacle collision detection in path search and dynamic constraints in DVS trajectory optimization, we use the minimum circumsphere of the formation shape to characterize the DVS as shown in Fig. \ref{frond}.
The state of the DVS includes the spatial coordinates of the centroid of the circumsphere ${p}_W$. To enable scalable swarms, provide sufficient swarm safety distances, and allow formation deformations for obstacle avoidance, the scaling factor governing the size of the virtual structure is integrated into the state space.
However, navigation through narrow environments often demands rapid adaptation of formation shape beyond simple scaling, as excessive uniform scaling could compromise swarm safety distances. To address this limitation, the state space of the virtual structure is further augmented to include degrees of affine transformations. Due to significant aerodynamic disturbances, particularly downwash effects inherent in drones, affine deformations are restricted in the horizon to ensure safety. We set the affine transformation parameter to 

\begin{align}
\boldsymbol{A} = \begin{bmatrix}
\alpha_{x} & 0 & 0 \\
0 & \alpha_{y} & 0 \\
0 & 0 & 1 
\end{bmatrix}, \alpha_{x} \cdot \alpha_y = 1.
\end{align}
By setting $\alpha_{x} \cdot \alpha_y = 1, \alpha=\alpha_x$, the affine transformation is characterized by a single parameter $\alpha$, which reduces computational complexity. Moreover, this can ensure that the area of the ellipse after affine deformation is the same as the original circle, enabling orderly agent arrangement and preventing excessive formation deformation.
Therefore, the expected position of the agent $i$ in formation with virtual body deformation is 
\begin{align}
p_i= {p}_W +
p_{i,c} \cdot \boldsymbol{A} \cdot \frac {\boldsymbol{R}} {\boldsymbol{R_0}},
\end{align}
where $\boldsymbol{R_0}^T = [r_0,r_0,1]$ indicates the initial radius vector of the virtual body. The $\boldsymbol{R}^T = [r, r, 1]$, and $r$ represents the scaling factor of DVS with deformation. To ensure safe swarm collision avoidance, we need to appropriately scale the desired formation. We calculate the minimum pairwise distance:
\begin{align}
l_{min} = \min \| p_{i,c}- p_{j,c}\|,
\forall i,j \in \{1,...,{n}\}, i \neq j.
\end{align} Based on the safety threshold $l_{s} =  2h\cdot r_a, h \geq 1$, the scaling factor $r=r_0 \cdot l_{s}/l_{min}$ is derived to proportionally adjust the target positions. $r_a$ is the radius of the actual agent.
$r_0$ denotes the radius of the DVS before the current PAAS.

\begin{figure}[!t]
    \centering
    \includegraphics[width=3.2in]{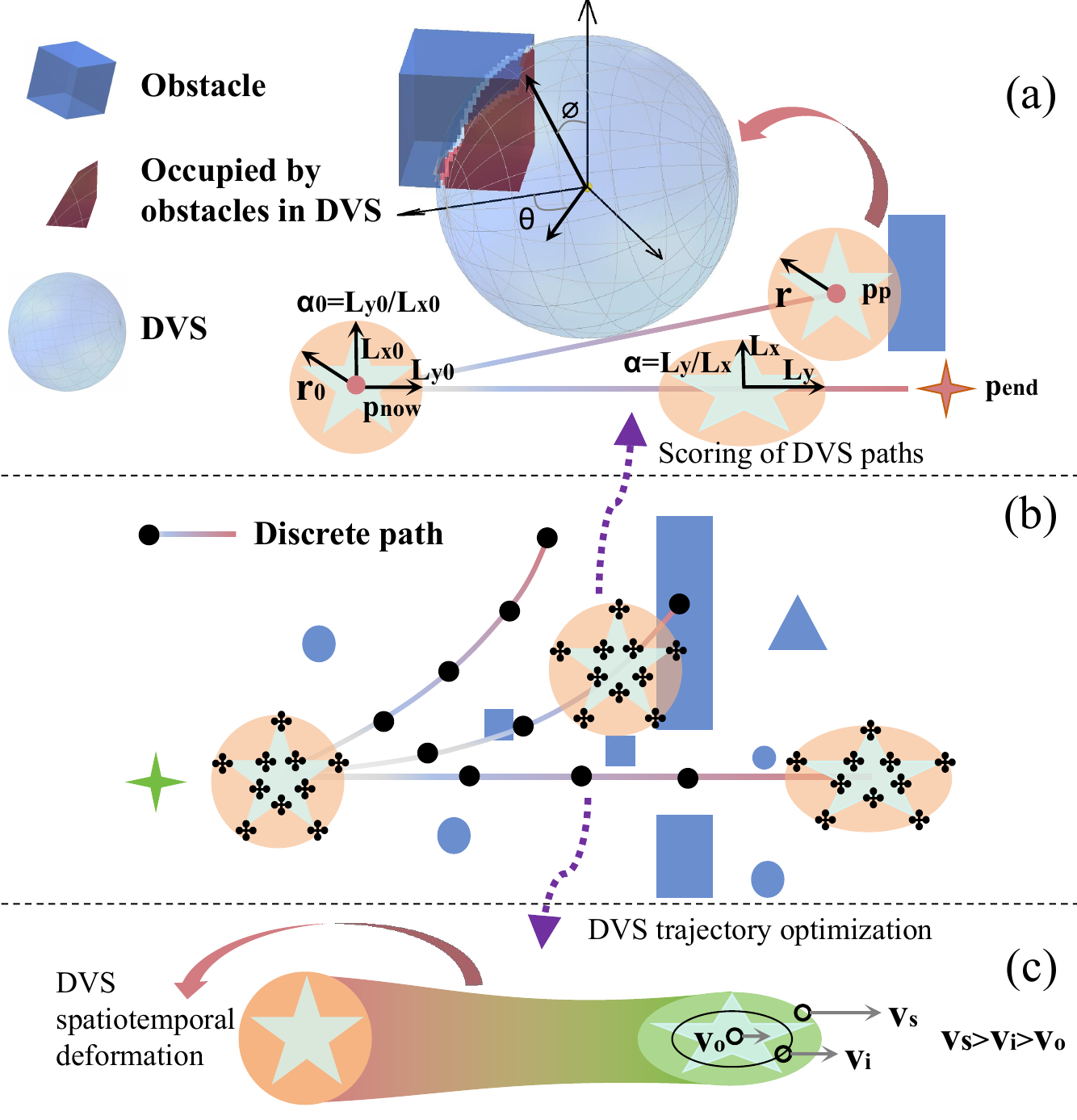}
    \vspace{-0.4cm}
    \caption{
    Schematic of DVS path scoring and DVS trajectory optimization based on the DVS path. (a) Key parameters for DVS path scoring in Eq. \ref{score_eq}. (b) Schematic of DVS path search. (c) The DVS path is used as an initial value to optimize DVS spatiotemporal trajectories, with dynamics constraints applied to points on the DVS outer contour to ensure dynamic feasibility of all points inside the DVS.
    }
    \vspace{-0.7cm}
    \label{score}
\end{figure}

We expand the state space of DVS to five dimensions, to accommodate the computational demands of real-time high-dimensional path search, we employ a primitive-based strategy for DVS path search. As illustrated in Fig. \ref{frond}, we store a library of candidate paths representing different spatial orientations of the virtual structure. Within each orientation, multiple path variants are stored with distinct scale and affine transformation configurations.
To mitigate computational complexity, we enforce uniform scale and affine parameters along a path. Although this design introduces a trade-off by potentially sacrificing path optimality, it significantly enhances the efficiency. As shown in Fig. \ref{score} (a), we define a scoring function to evaluate path quality:
\begin{align}
S &=1/(w_r \cdot c_r+w_{\alpha} \cdot c_{\alpha}+w_o \cdot c_o+w_e \cdot c_e), \nonumber \\
& c_r =||r-r_0||^2, \nonumber\\
& c_{\alpha} = ||\alpha-\alpha_0||^2, \nonumber\\
& c_o = {\frac{3}{4\pi r^3}}\int_{0}^{2\pi}\int_{0}^{\pi}\int_{0}^{r} O({p}_{Wi} +[p_x,p_y,p_z])  dr d\theta d\phi, \nonumber\\
& c_e = \frac{(p_{end}-p_{now})\cdot (p_{p}-p_{now})}{\| p_{end}-p_{now}\|\cdot \| p_{p}-p_{now}\|}, \label{score_eq}
\end{align}

\begin{equation}
\begin{aligned}
& O(x,y,z)= \left\{
      \begin{aligned}
      &0, &{\rm if} \, point \ (x,y,z) \in \mathcal{F},\\
      &1,      &{\rm otherwise},\\
      \end{aligned}
      \right.
\end{aligned}
\end{equation}
where $c_r$ and $c_{\alpha}$ which represent the difference from the initial $r$ and ${\alpha}$ are set to reduce unnecessary deformation, respectively. $r_0$ indicates the initial scaling factor of the virtual body, $\alpha_0$ represents the initial affine transformation factor. $c_o$ represents the obstacle occupancy ratio within the virtual structure, which can be computed by discretizing the spherical space and performing cumulative sum in practical algorithmic implementations. It leverages the tolerance of formation for internal obstacles during path search. Where $p_x = \alpha \cdot r \cdot cos(\phi)$, $p_y = r\cdot sin(\phi)/\alpha$, $p_z = r\cdot sin(\theta)$, $\theta$ is the polar angle and $\phi$ is the azimuthal angle in spherical coordinates, $\mathcal{F}$ represents obstacle-free space. $c_e$ represents the degree to which the path endpoint direction aligns with the formation's target destination. $p_{now}, p_{end}, p_{p}$ are the current DVS center position, DVS endpoint position, and primitive path endpoint position, respectively.
% , and $C_3$ represents the obstacle occupancy ratio within the virtual structure

\subsection{Trajectory Optimization of Deformable Guidance}
\label{mid_end}
To ensure the feasibility of agent dynamics, we incorporate the control effort of the DVS scale and affine transformations to maintain stable formation deformation. In addition, the flight time cost is introduced. 
In this paper, The trajectory of the DVS adopts MINCO \cite{c32} as the trajectory representation. We set $\sigma(t)$ to represent an $m$-dimensional $M$-piece polynomial trajectory of degree $N = 2s-1$ and $s$ denotes the order of the relevant integrator chain. For the trajectory of the DVS, $\bm \sigma(t)=[{p}^T_W,r,\alpha]^T(t)$. The polynomial coefficient $\textbf{c} = (\textbf{c}^T_1, ..., \textbf{c}^T_M)^T \in \mathbb{R}^{2Ms \times m}$ is derived from $\mathcal{M}(\textbf{q},\textbf{T})$, where $\textbf{q} = (\textbf{q}_1, ..., \textbf{q}_{M-1})$ represent waypoints, $\textbf{T} = (T_1, T_2, ..., T_M)^T$ represents the time allocated for each trajectory segment. $T_\Sigma = \sum_{i=1}^M T_i$ signifies the total time span. 
Each $m$-dimensional $M$-piece trajectory is defined as
\begin{align}
\bm \sigma(t) = \bm \sigma_i(t - t_{i-1}) \quad \text{for all } t \in [t_{i-1}, t_i),
\end{align}
we use a 5 order polynomial, and the $i^{th}$ piece trajectory is
\begin{align}
\bm \sigma_i(t) = \textbf{c}^T_i \beta(t) \quad \text{for all } t \in [0, T_i),
\end{align}
where $\textbf{c}_i \in \mathbb{R}^{(N+1) \times m}$ represents the coefficient matrix, $\beta(t) = [1, t, ..., t^N]^T$ is the natural basis and $T_i = t_i - t_{i-1}$ is the time allocated for the $i^{th}$ piece.
The trajectory planning problem is formulated as a non-linear problem with the constraint of guaranteed dynamic feasibility of all points inside the DVS:
\begin{align}
	\min_{\mathbf{c}, \bm T} J &= 
	\int_{0}^{T} {p}_W^{(s)}(t)^{\rm T} {p}_W^{(s)}(t) 
	+ ({r}^{(s)}(t))^2 \nonumber \\
	&+ (\alpha^{(s)}(t))^2 + w_TT \\
	s.t. & \bm \sigma^{[s-1]}(0) = \bm \sigma_0, \bm \sigma^{[s-1]}(T) = \bm \sigma_f \label{eq1},
	\\
	&||{p}_{j,s}^{(1)}(t)||_2^2 \leq {v}_{\max}^2, ||{p}_{j,s}^{(2)}(t)||_2^2 \leq {a}_{\max}^2, \nonumber \\
       &\forall j \in \{1,...,{N_f}\} , \label{eq2}\\
	&\bm \sigma^{[\widetilde{d}]}_i(T_i) = \bm \sigma^{[\widetilde{d}]}_{i+1}(0), \label{eq3}
	\\
	&\forall t \in [0,T], \forall i \in \{1,...,{M}\}. \label{eq4}
\end{align}
Here, $\boldsymbol{p}_{j,s}(t)$ is the $j^{th}$ point of the outer contour in DVS at time $t$ and $N_f$ is the number of $\boldsymbol{p}_{j,s}(t)$, while
\begin{align}
&{p}_{j,s}(t) = {p}_W(t) + {p}_{j,s_0} \cdot \boldsymbol{A}(t) \cdot \frac {\boldsymbol{R}(t)}{\boldsymbol{R_0}}, \\
 &{p}_{j,s_0} =  \{ r cos \frac{a\theta}{N_\theta}cos \frac{b\phi}{N_\phi}, r cos \frac{\theta}{N_\theta}sin\frac{b\phi}{N_\phi}, r sin \frac{\theta}{N_\theta} \}, \nonumber \\ & \forall a \in \{0,...,{N_{\theta}}\}, \forall b \in \{0,...,{N_{\phi}}\},
\end{align}
where ${p}_{j,s_0}$ is the $j^{th}$ discretized sampling of the surface point within the initial DVS shape. $N_{\theta}$ and $N_{\phi}$ are the discretization resolution alone the polar angle and the azimuthal angle, respectively. As shown in Fig. \ref{score} (c), it can be inferred that when the dynamics of boundary points remain within their limits, the internal points of the DVS will also satisfy the dynamic constraints. Eq. \ref{eq1} ensures the state constraints at both the start and end points of the DVS trajectory, while Eq. \ref{eq2} enforces that all internal points along the trajectory meet predefined thresholds for the velocity and acceleration limits. Furthermore, Eq. \ref{eq3} imposes continuity constraints at the waypoints connecting adjacent polynomial segments to ensure smooth transitions. 

The equality constraints Eq. \ref{eq1} \& Eq. \ref{eq3} are eliminated using MINCO variables, enabling efficient constrained optimization in real-time applications. For handling inequality constraints Eq. \ref{eq2}, we adopt a penalization strategy \cite{penopt}, where constraint elimination and optimality assurance follow the methodology outlined in \cite{c32}. This framework ultimately converts the original constrained formulation into an unconstrained optimization problem, and we use L-BFGS \cite{lbfgs} to solve it efficiently.

% \begin{equation}
% \rm P3: {\rm min} \, [\mathcal J_f, \mathcal J_o] \cdot \mu,
% \end{equation}

% where $\mu$ denotes the weight vector. $\mathcal J_o$ contains the cost function of the control effort, flight time, collision avoidance, and dynamical feasibility. $\mathcal J_f=f(\mathcal F_f)$ represents the cost function of similarity formation, and $f(\cdot)$ is the differentiable metric to quantify the similarity distance $\mathcal F_f=||\mathbf L_{mc}^{sqr}- {(\mathbf{L}^{sqr}_{mc})}_{des}||_F^2$ between current and desired formations. $\mathbf L_{mc}^{sqr}$ is the Laplacian matrix that represents the sparse graph $\mathcal G_{mc}^{spr}$, and ${(\mathbf{L}^{sqr}_{mc})}_{des}$ denotes the matrix describing the desired formation configuration in the maximum clique. $||\cdot||_F$ is Frobenius norm.

\subsection{Agent Trajectory Optimization Guide by DG}
\label{back_end}
After obtaining the optimized formation guidance trajectory of the DVS and the target positions, they will be broadcast to all other agents for formation cooperation. We also employ MINCO in \ref{mid_end} for agent trajectory parameterization and $\bm \sigma(t)={p}^T(t)$ for the agent. We formulate agent trajectory planning as an optimization problem:

\begin{align}
& \min_{\textbf{q},\textbf{T}} \quad \int^{t_M}_{t_0} {\| p^{(s)}(t) \|}^{2}dt\ + \ \rho \cdot{T_\Sigma} , \\
& \text{s.t.} \quad p(t) = \mathcal{M}_{\textbf{q},\textbf{T}} \quad \forall t \in [t_0,t_M], \\
& \quad \quad \mathbf{p}^{[s-1]}(0) = \bar{\mathbf{p}}_0,\\
& \quad \quad \mathbf{p}^{[s-1]}(t_M) = \bar{\mathbf{p}}_f ,\\
& \quad \quad \mathcal{H}(p(t), ...,p^{(s)}(t)) \preceq 0 \quad \forall t \in [t_0,t_M] ,
\end{align}

The trajectory $p(t)$ is defined by the optimization variables $\{ \textbf{q},\textbf{T}\}$ and the temporal regularization coefficient $\rho$. The state vector $\mathbf{p}^{[s-1]}(t) = (p(t)^T,\dot p(t)^T,...,p^{[s-1]}(t)^T)^T\in \mathbb R^{ms}$ encodes the $[s-1]^{th}$ order derivatives of a chained dynamical system governed by $s$-integrator dynamics. The boundary conditions are specified by the initial state $\bar {\mathbf{p}}_0$ and the terminal state $\bar {\mathbf{p}}_f$. 

The continuum constraint set $\mathcal{H}$ encompasses dynamic feasibility $\mathcal{J}_d$, obstacle clearance $\mathcal{J}_o$, and swarm reciprocal avoidance $\mathcal{J}_s$. Those constraints are modeled by using the same method in \cite{c1}, the reader can refer to \cite{c1} for details. For formation coordination $\mathcal{J}_f$, we use the DVS trajectory as formation coordination constraints. Specifically, based on the desired relative position assigned to agent $i$ within the formation, we can derive the spatiotemporal desired formation position sequence $\mathbf{p}_{i,f} = \{ {p}_{i,f}(0),\dots,{p}_{i,f}(j),\dots, {p}_{i,f}(M_c)\}$ on the DVS spatiotemporal trajectory over a future time period with a sequence of time stamps $\{0,\dots,j,\dots, M_c\}$. This establishes the formation constraint for the agent $i$ as follow
\begin{align}
\mathcal{J}_f \implies \sum_{j=0}^{M_c} \| p(j) - {p}_{i,f}(j)\|^2.
\end{align}
By employing the same constraint elimination method described in Sec. III-B, we also transform the constrained optimization problem into an unconstrained optimization problem which can be solved by L-BFGS.

\section{Benchmarks and Experiments}
\label{sec:experiments}

Simulation experiments, benchmarks and real-world experiments are conducted to validate the effectiveness and superiority of our methods. The simulation platform is a personal computer with an Intel Core i7 8700K CPU running at 3.2 GHz and with 32-GB RAM at 3200 MHz.

% \begin{figure}[!t]
%     \centering
%     \includegraphics[width=3.5in]{pics/sim2.pdf}
%     \vspace{-0.3cm}
%     \caption{
%     A heart shape formation planning in the obstacle environment with varying number of drones.(a) Initial formation of 20 agents with 4 agents and 6 agents awaiting join in the formation, respectively. (b) Formation flight to the destination with agents trajectory visualization. (c) Snapshot of the initial formation. (d)-(e) Snapshots of 4 agents joining the formation for the first time and followed 6 joining for the second time. (f)-(g) Snapshots of 3 agents leaving the formation for the first time and followed 4 leaving for the second time.
%     }
%     \vspace{-0.6cm}
%     \label{xin}
% \end{figure}

\begin{table}[thpb]
  \renewcommand{\arraystretch}{1.3}
  \caption{Calculation time of PAAS for different shapes}
  \label{alas_tab}
  %\footnotesize
  \setlength\tabcolsep{6.5pt} %space between two columns.
  \centering
  \begin{threeparttable}
    \begin{tabular}{cccccc}%{\textwidth}
      \toprule
      \textbf{Number of agents} & {20} & {40} & {60} & {80} & {100}\\
      \hline
      \textbf{formation shape} &\makecell[c]{\emph{$ t_{pa}(s)$}} &\makecell[c]{\emph{$ t_{pa}(s)$}} &\makecell[c]{\emph{$ t_{pa}(s)$}} &\makecell[c]{\emph{$ t_{pa}(s)$}} &\makecell[c]{\emph{$ t_{pa}(s)$}} \\
      \hline
      \makecell[c]{\emph{Equilateral triangle}} &\makecell[c]{0.0374} &\makecell[c]{0.0590} &\makecell[c]{0.0823} &\makecell[c]{0.1029} &\makecell[c]{0.1307} \\
      \hline
      \makecell[c]{\emph{Square}} &\makecell[c]{0.0419} &\makecell[c]{0.0575} &\makecell[c]{0.0833} &\makecell[c]{0.1034} &\makecell[c]{0.1230} \\
      \hline
      \makecell[c]{\emph{Pentagram}} &\makecell[c]{0.0511} &\makecell[c]{0.0586} &\makecell[c]{0.0864} &\makecell[c]{0.0936} &\makecell[c]{0.1105} \\
      \hline
      \makecell[c]{\emph{Heart shape}} &\makecell[c]{0.0588} &\makecell[c]{0.0827} &\makecell[c]{0.1138} &\makecell[c]{0.1345} &\makecell[c]{0.1607} \\
      \bottomrule
    \end{tabular}
    \begin{tablenotes}
      \item \footnotesize $t_{pa}$: The maximum calculation time of PAAS in multiple tests; 
    \end{tablenotes}
    \vspace{-0.6cm}
  \end{threeparttable}
\end{table}

\begin{figure*}[!t]
    \centering
    \vspace{-0.6cm}
    \includegraphics[width=6.5in]{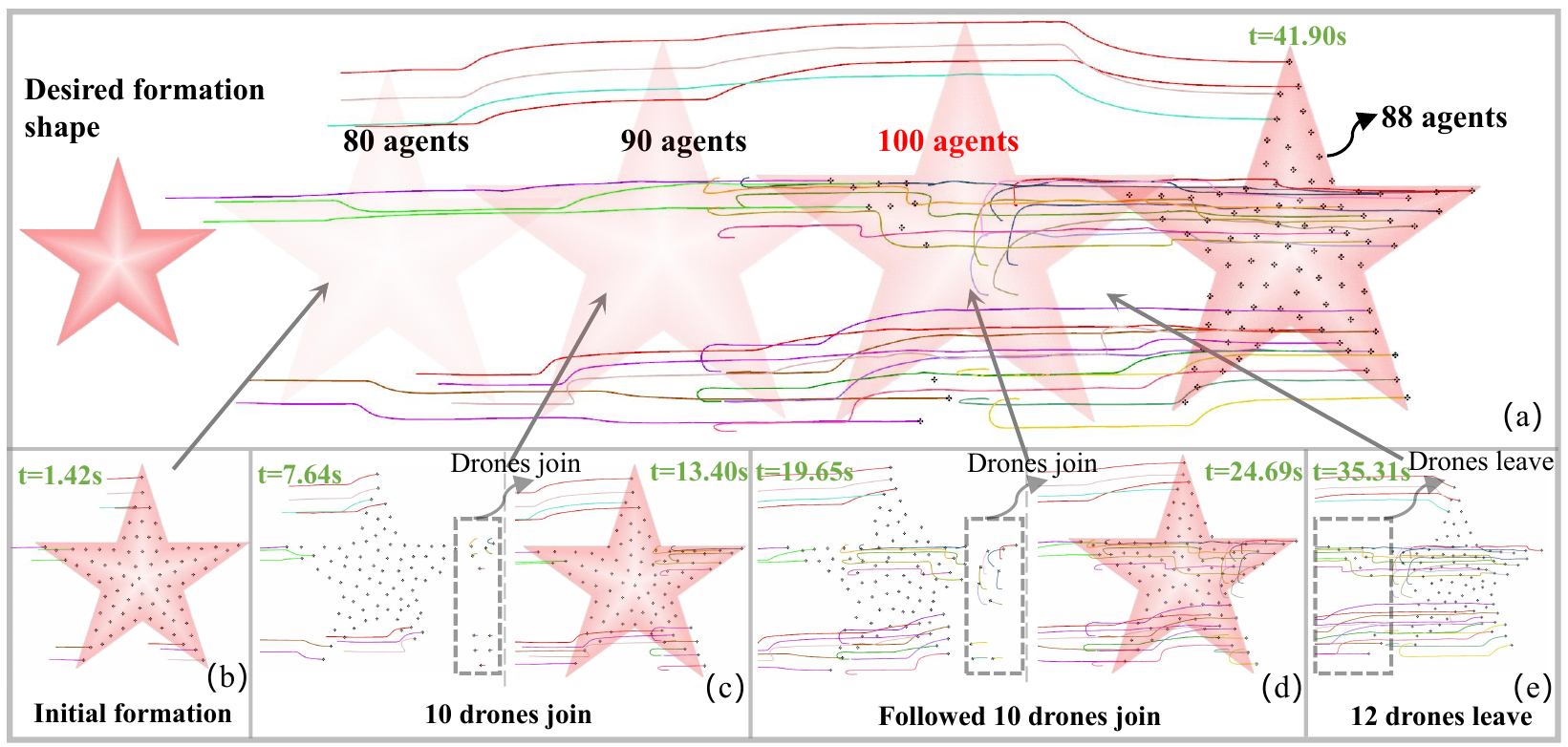}
    \vspace{-0.4cm}
    \caption{
    A pentagram formation flight with varying number of drones in large-scale swarm. (a) Formation recovery and motion process with varying swarm numbers during flight. (b) Snapshot of the initial formation with 80 agents. (c) Process of 10 agents joining the formation. (d) Process of 10 agents joining the formation for the second time. (e) Snapshot of 12 agents leaving the formation.
    }
    \vspace{-0.3cm}
    \label{star}
\end{figure*}

\begin{figure*}[!t]
    \centering
    \includegraphics[width=6.5in]{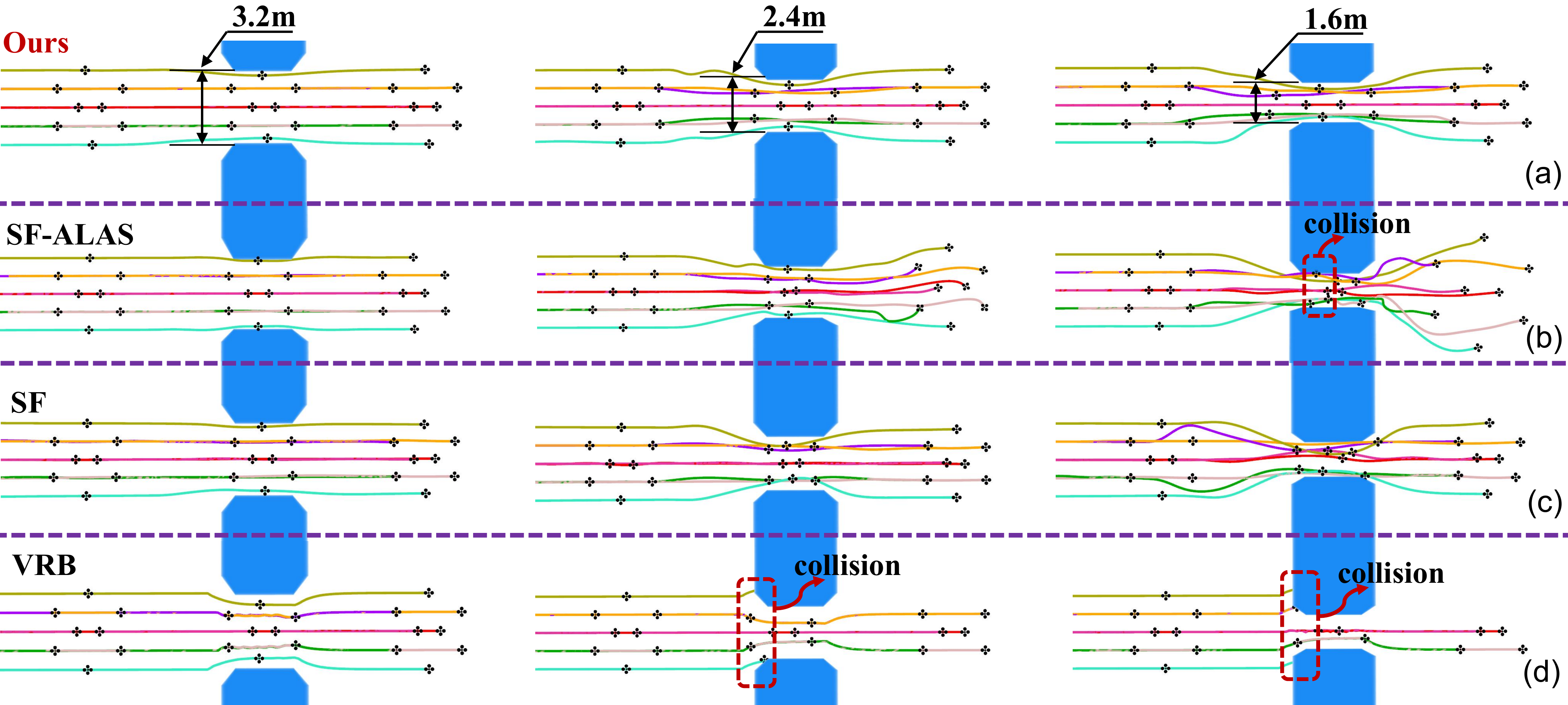}
    \vspace{-0.3cm}
    \caption{
    Benchmarks on narrow gap traversal with comparative methods. (a)-(d) Representative snapshots of Ours, SF-ALAS, SF, and VRB traversing gaps with widths of 3.2$m$, 2.4$m$, and 1.8$m$, respectively.
    }
    \label{bench}
    \vspace{-0.3cm}
\end{figure*}

\begin{table}[thpb]
  \renewcommand{\arraystretch}{1.3}
  \caption{Different methods for benchmarks}
  \label{methods}
  %\footnotesize
  \setlength\tabcolsep{6.0pt} %space between two columns.
  \centering
  \begin{threeparttable}
    \begin{tabular}{cccccc}%{\textwidth}
      \toprule
      \textbf{Methods} &\makecell[c]{Description} \\
      \hline
      \makecell[c]{{Ours}} &\makecell[c]{Formation planning guided by DG} \\
      \hline
      \makecell[c]{{SF} \cite{tro_icra}} &\makecell[c]{Distributed Swarm Formation planning} \\
      \hline
      \makecell[c]{{SF-ALAS}\cite{c1}} &\makecell[c]{Swarm reorganization by ALignment and ASsignment} \\
      \hline
       \makecell[c]{{VRB}\cite{vrb}} &\makecell[c]{Formation planning guided by rigid virtual Structures} \\
      \bottomrule
    \end{tabular}
    \vspace{-0.4cm}
  \end{threeparttable}
\end{table}

\begin{table*}[thpb]
  \renewcommand{\arraystretch}{1.3}
  \vspace{-0.6cm}
  \caption{Benchmarks in narrow environment}
  \label{bench_table}
  %\footnotesize
  \setlength\tabcolsep{15.0pt} %space between two columns.
  \centering
  \begin{threeparttable}
    \begin{tabular}{cccccc}%{\textwidth}
      \toprule
      \textbf{Width of gap} \emph(m) & {3.2} & {2.8} & {2.4} & {2.0} & {1.6}\\
      \hline
      \makecell[c]{Methods} &\emph{$r_{s}$(\%)} \emph{ } \emph{$t_{f}$(s)} &\emph{$r_{s}$(\%)} \emph{ } \emph{$t_{f}$(s)} &\emph{$r_{s}$(\%)} \emph{ } \emph{$t_{f}$(s)} &\emph{$r_{s}$(\%)} \emph{ } \emph{$t_{f}$(s)} &\emph{$r_{s}$(\%)} \emph{ } \emph{$t_{f}$(s)} \\
      \hline
      \makecell[c]{{Ours}} &\makecell[c]{100} \makecell[c]{7.41} &\makecell[c]{100} \makecell[c]{7.43} &\makecell[c]{100} \makecell[c]{7.39} &\makecell[c]{100} \makecell[c]{7.42} &\makecell[c]{\textbf{100}} \makecell[c]{\textbf{7.48}} \\
      \makecell[c]{{SF}} &\makecell[c]{100} \makecell[c]{7.99} &\makecell[c]{100} \makecell[c]{8.61} &\makecell[c]{100} \makecell[c]{8.59} &\makecell[c]{100} \makecell[c]{8.30} &\makecell[c]{100} \makecell[c]{8.35} \\
      \makecell[c]{{SF-ALAS}} &\makecell[c]{100} \makecell[c]{7.11} &\makecell[c]{94} \makecell[c]{7.15} &\makecell[c]{86} \makecell[c]{7.28} &\makecell[c]{30} \makecell[c]{7.30} &\makecell[c]{26} \makecell[c]{7.33} \\
      \makecell[c]{{VRB}} &\makecell[c]{100} \makecell[c]{7.07} &\makecell[c]{0} \makecell[c]{N/A} &\makecell[c]{0} \makecell[c]{N/A} &\makecell[c]{0} \makecell[c]{N/A} &\makecell[c]{0} \makecell[c]{N/A} \\
      \bottomrule
    \end{tabular}
    \begin{tablenotes}
      \item \footnotesize $r_{s}$: Success rate (\%); $t_{f}$: Flight time (s); N/A: Planning fail, i.e., collision during flight.
    \end{tablenotes}
    \vspace{-0.5cm}
  \end{threeparttable}
\end{table*}

\begin{figure*}[!t]
    \centering
    \includegraphics[width=6.5in]{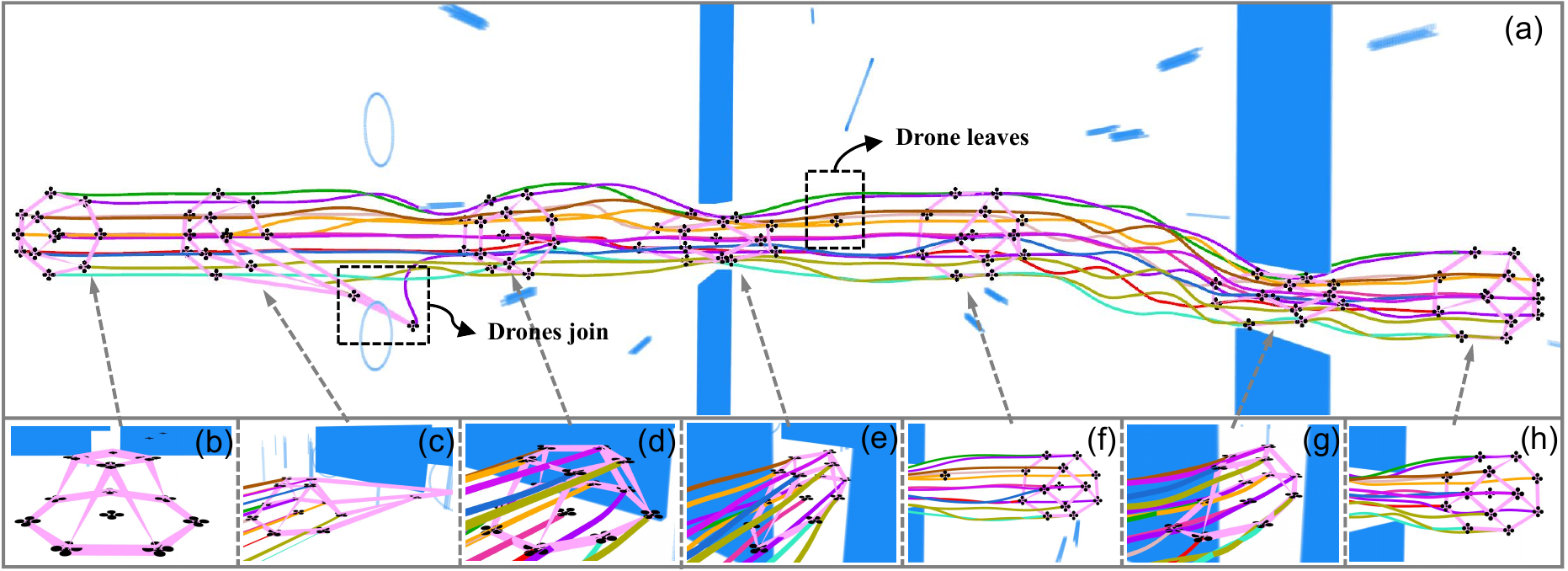}
    \vspace{-0.3cm}
    \caption{
    Simulation in narrow environments with varying number of drones. (a) Process of the simulation on the formation flight with 14 agents, including number changes and continuously pass through two narrow gaps guided by DG. (b)-(h) Snapshots of the key transition stages of formation during the entire flight process.
    }
    \vspace{-0.5cm}
    \label{all}
\end{figure*}

\begin{figure*}[!t]
    \centering
    \vspace{-0.7cm}
    \includegraphics[width=6.5in]{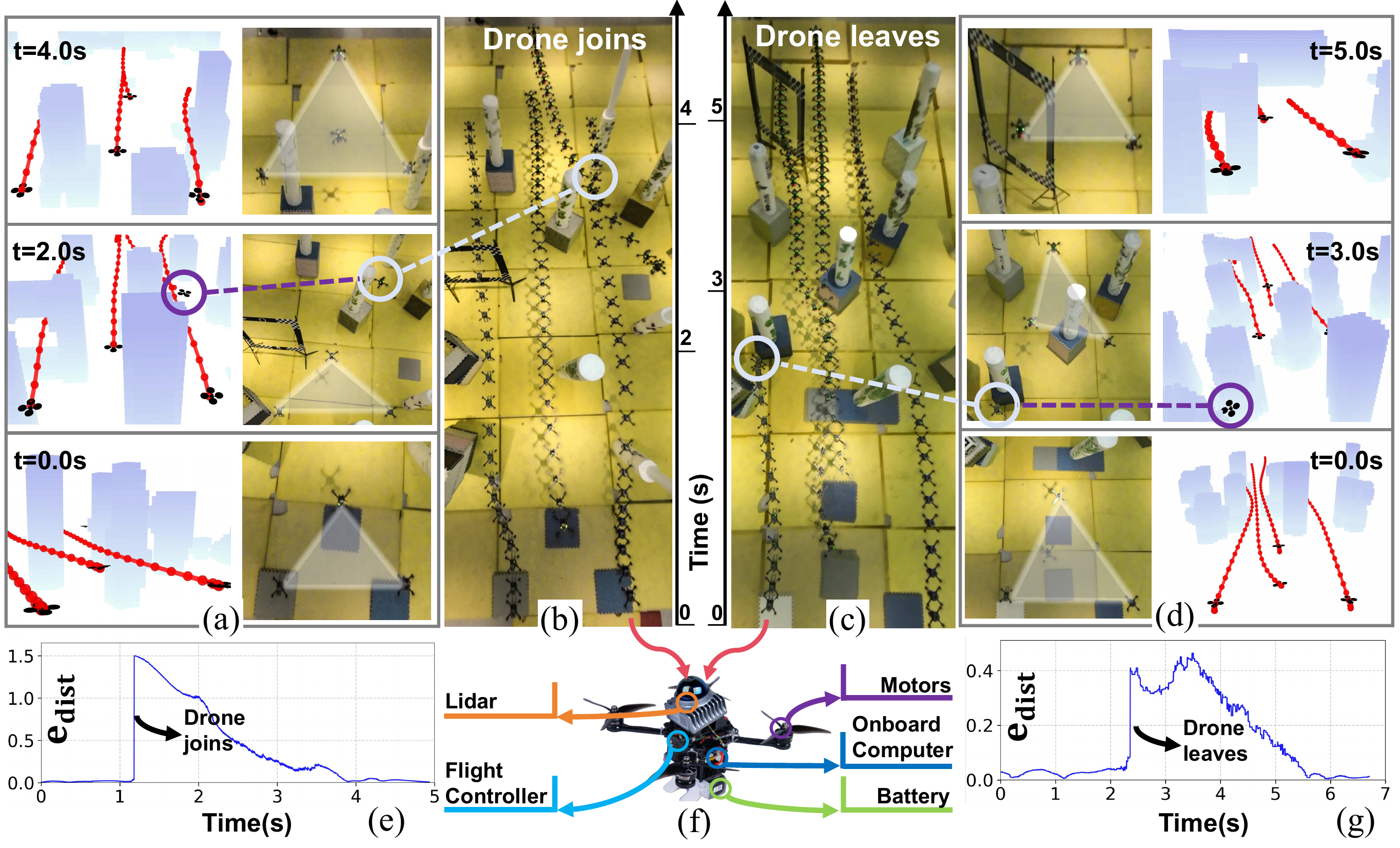}
    \vspace{-0.4cm}
    \caption{
    Real-world experiments of agents join and leave. (a) Screenshots and Rviz diagrams of corresponding formations at different times for agent joins. (b) The process of an agent joining a formation of three agents, while the formation is traversing an obstacle area guided by DG. (c) The process of an agent leaving a formation of four agents. (d) Screenshots and Rviz diagrams of corresponding formations at different times for agent leaves. (e) Formation error $e_{dist}$ in the flight process of agent join. (f) Quadrotor in real world experiments. (g) Formation error $e_{dist}$ in the flight process of agent leaving.
    }
    \vspace{-0.3cm}
    \label{real_2}
\end{figure*}

\begin{figure}[!t]
    \centering
    \includegraphics[width=3.3in]{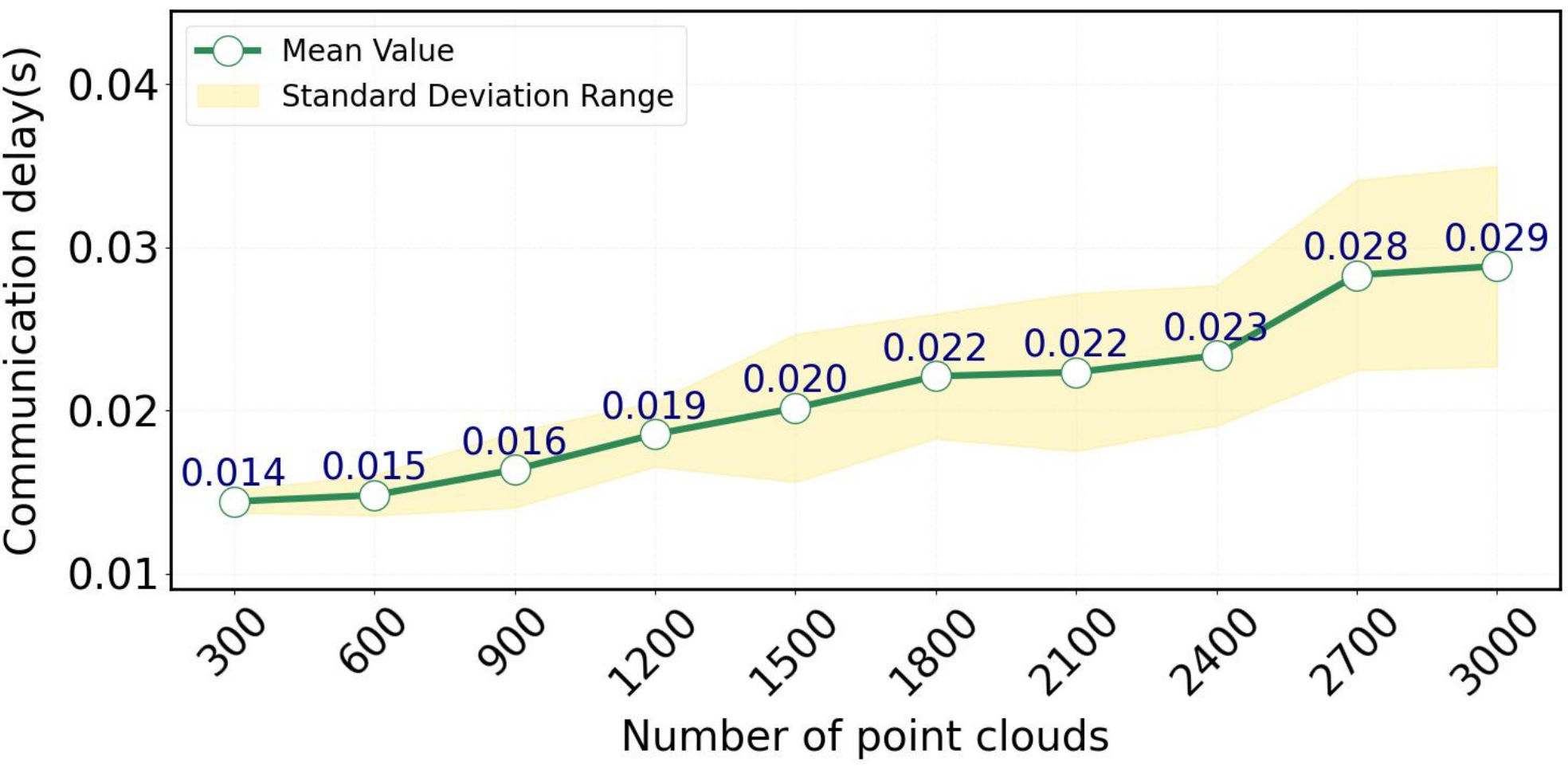}
    \vspace{-0.4cm}
    \caption{
    Average and standard deviation of point cloud communication delay versus number of points per communication.
    }
    \vspace{-0.5cm}
    \label{point}
\end{figure}

\subsection{Simulation Experiments under Varying Number}
\label{sim_exp}
% 我们的方法可以实现在复杂环境中当集群数量变化时编队能够根据期望编队图形进行均匀位置划分以及分配恢复成新编队。为了说明我们方法的有效性，我们设置了存在部分智能体连续多次加入和退出整体编队的仿真实验。在本文中我们不关注飞机进入与失效的检测，每一次变化都是外部主动发送给leader的增加与退出的消息。如图1所示，我们选择爱心图形这种非凸图形作为编队期望图形，初始编队为20个无人机。编队运动前期依次加入4个智能体以及6个智能体，最大加入智能体的比例为变化前的集群数量的25%。编队运动后期依次退出3个智能体以及4个智能体，最大退出智能体的比例为变化前的集群数量的15%。如图所示，系统能够根据集群数量以及期望编队形状快速划分新的期望位置以及进行位置分配，同时依靠DG在障碍物环境中快速恢复成期望编队。
Our method enables the swarm to achieve uniform position partitioning and assignment based on the desired formation shape when the number of agents changes. To validate effectiveness, we conduct simulation in which agents dynamically join and leave the swarm multiple times. In simulations, the upper limit for agent velocity is set to $2.0 m/s$, and the upper limit for acceleration is set to $7 m^2/s$. We do not address agent entry and failure detection, and each change event is triggered by external commands actively sent to the DG. As shown in Fig. \ref{fig1}(d)\&(e), a non-convex heart-shaped pattern is chosen as the target formation shape, with an initial swarm of 20 agents. During the early movement phase, 4 agents join the swarm, followed by 6 agents that reach 25\% of the current swarm number. In the later phase, 3 agents leave the swarm, followed by 4 agents that reach 15\% of the current swarm number. The system rapidly partitions target positions based on updated swarm number and desired shape. By leveraging DG, it achieves obstacle avoiding convergence to the new formation, demonstrating efficient recovery capability.

% 为了额外说明我们重新划分与分配能力的效率，同时说明我们方法在大规模上编队引导能力，我们进一步部署了一个100飞机的具有无人机加入退出的实验。我们在这个100无人机实验中没有考虑障碍物环境，而是更关注编队保持以及划分与分配效率。
% 如图1所示，我们设置编队期望形状为五角星形状，初始集群数量为80个智能体，这80个智能体提前使用均匀划分来获得期望编队位置。t=1s时候，另外的10个智能体加入到编队当中，DGE重新自主计算划分与分配， 并且在DG的引导下进行轨迹优化从而快速恢复成期望编队形状以及进行后续稳定的编队保持。编队中的智能体在1s内生成同样形状的编队，同时为了保证机间安全，整体的形状scale扩大到初始编队的1.2倍。t=2秒时候，另外的10个智能体再次加入到编队当中，同样在1s内快速形成期望编队形状，同时整体的形状scale适应性的增大到初始编队的1.4倍。最后在t=3s的时候，编队中的12个智能体退出编队协同，其他智能体在1s内重新形成期望编队形状， 同时整体的形状scale适应性的减少到编队在t=2s时的scale的0.7倍。在整个过程中，我们的系统支持最高100智能体的实时划分与分配，如tab1所示，我们实验测试了不同数量下的划分与分配的计算耗时（列举计算机设备配置），每个数量的测试都测试了20次，最后计算计算耗时的平均值。可见我们的方法在智能体数量为100时候计算耗时为1s，完全满足大规模编队规划实时计算的需求。

% 最后我们部署了在复杂环境中一个包含编队数量变化以及DVS仿射变形引导编队连续穿越两个窄缝的综合的仿真实验。期望编队为一个圆台，首先划分圆台为俩层，根据两层切片面积比例，初始编队划分为上下两层一共12个智能体。在t=1s时候加入了2个智能体，通过面积比例划分两层智能体比例，再经过划分和分配形成新的14个智能体的编队，然后经过DVS的变形引导陆续穿过两个窄缝，其中两个窄缝中心不在同一水平线上。总而言之我们的系统能够在复杂障碍物环境中应对数量变化的编队规划问题，实时进行新编队位置的划分与分配，并且快速恢复成期望编队穿越复杂障碍物环境。
To further demonstrate the efficiency of our PAAS capabilities while validating the scalability of our method, we deploy an additional experiment with a large-scale swarm involving agents joining and leaving events. We deliberately exclude obstacle environments to focus specifically on the effectiveness of PAAS. As shown in Fig. \ref{star}, we set the desired formation shape as a pentagram, with an initial swarm number of 80. These 80 agents obtain their desired formation positions by using PAAS previously. At $t=17.64s$, 10 agents join the formation and the DG executes the PAAS. Then, the trajectory optimization is performed to quickly restore the desired formation shape and maintain stable formation under the guidance of DG. The agents in the formation generate a larger pentagram in about $6.0$ seconds. To ensure swarm safety, the system scales the formation geometry to about $115 \%$ of the original size. At $t = 19.65s$, 10 additional agents integrate into the formation. Our method achieves the target configuration in about $5.0$ seconds, with an autonomous scaling of the formation geometry to about $125 \%$ of the original size. Finally, at $t=35.31s$, 12 agents leave this formation and the remaining agents reform the desired formation shape in about $7.0$ seconds, with the overall shape scale adaptively reduced to $85 \%$ of the formation size at $t=41.90s$. Throughout the process, our system supports real-time PAAS for up to 100 agents. As shown in the Tab. \ref{alas_tab}, we experimentally test the computation time for PAAS in different numbers. Each case is tested 20 times, and the average computation time is provided. It can be seen that our method costs $0.1592 s$ when the number of agents reaches 100. Given the agent’s maximum speed limit of $2 m/s$, the maximum movement distance within $0.2 s$ is constrained to $0.4 m$, which falls within its perception range. This permits DG to optimize the DVS trajectories for guiding agents to converge to the desired formation and meets the requirements of real-time large-scale formation planning.

\subsection{Benchmarks in Narrow Environment}
\label{benchmark}

To validate the superior efficiency of obstacle negotiation in narrow environments, we implement a comparative traversal test through narrow gaps. As shown in Tab. \ref{methods}, we conduct comprehensive simulations comparing three state-of-the-art methods, including a high performance formation planner for narrow environments (SF) \cite{tro_icra}, a formation planning method that can adjust the general deformation of the formation by reconfiguring the local formation targe position (SF-ALAS) \cite{c1}, and a formation planning method guided by rigid virtual structure (VRB) \cite{vrb}. To ensure fairness, all test environments, initial formation positions, and kinematic constraints are standardized across methods. Crucially, as comparative methods lack autonomous position repartitioning capabilities for number adaptation, there is no number change of drones in this benchmark.

As shown in Tab. \ref{bench_table}, we design five gaps of varying width. Each method is tested 50 times for traversing each gap, recording both the collision-free success rate and the traversal flight time. The flight time was calculated with the starting point set at 7$m$ to the left of the center of the gap, and the endpoint is determined when the formation error fell below 0.05 at 7$m$ to the right of the center of the gap. We adopt the formation error calculation method demonstrated in our previous work \cite{sparse}. As illustrated in Fig. \ref{bench}, we present a representative planning outcome for gravitational configurations of different widths $w_g$=3.2$m$, 2.4$m$, and 1.6$m$. Our method leverages the DVS environmental-adaptive deformation capability to alleviate formation pressure when avoiding obstacles, while ensure that affine transformations do not compromise swarm safety. Across 50 trials, our method achieves collision-free traversal through all gaps with consistently smooth agent trajectories that exhibit the lowest collision risk. In addition, the traversal times remained stable regardless of gap width variations. The SF-ALAS method attempts to resolve conflicts between formation maintenance and obstacle avoidance in narrow gaps through the single readjustment of local formation endpoints. When confronting narrow gaps, it frequently triggers multiple failed adjustments, leading to oscillatory trajectories. Such oscillations exacerbate collision risks as the gap width decreases, resulting in significantly low planning success rates. The success rate decreases to 94\% when $w_g$=2.8$m$, and the method experiences performance degradation at 2.0$m$ where success rates plummet to 30 percent, revealing limitations in a narrow environment. Endpoint adjustment alleviates collision avoidance pressure during gap traversal, facilitating convergence in this multi-objective optimization problem. This optimization yields time-efficient trajectories and achieves faster traversal times. As the width of the gap decreases, the increasing difficulty of adjustment extends the duration of the traversal. The method can mitigate collision avoidance pressure through adjustment when the gap widths below 2.0$m$. However, excessive scaling compromises swarm safety, making formation maintenance impossible during narrow-gap traversal.

The SF method, which operates without global adjustment mechanisms, demonstrates stable and safe traversal across gaps of different widths due to the absence of sudden changes in formation coordination conditions like SF-ALAS. However, when the gap width falls below 2.4$m$, this method does not maintain the integrity of the formation, showing severe oscillations in the flight trajectory. In the multi-objective optimization framework, the absence of adjustment mechanisms leads to excessive penalty costs for both obstacle avoidance and formation maintenance. This results in suboptimal solutions that compromise trajectory optimality, consequently yielding longer traversal times compared to alternative approaches. VRB represents an advanced rigid virtual body method. It demonstrates robust formation maintenance capabilities in obstacle-free environments, but fails to adapt to confined spaces. The formation successfully navigates obstacles when g=3.2$m$. Since agents only need to coordinate with the virtual body's movement, which maintains near-linear motion without environmental deformation, VRB achieves a significantly lower traversal time. However, the VRB method proves almost entirely ineffective, with no successful gap traversal recorded in narrower gaps.

In general, our method leverages DVS-guided spatiotemporal deformation to achieve superior performance in narrow environments. Compared to alternative methods, it demonstrates higher success rates in narrow-gap traversal, faster formation transit speeds, smoother collective trajectories.

% 结果现实我们在狭窄通道与大遮挡障碍物的复杂环境中的强大安全通过能力，我们的编队运行时间最短，成功率最高。
\subsection{Simulation in Narrow Environment with Varying Number}
We further deploy a comprehensive simulation of our method that involves both swarm number changes and the narrow environment. Our method guided by DG continuously passes through two narrow gaps. As shown in Fig. \ref{all}, the desired formation is a frustum which is divided into two layers. Based on the area ratio of the two layers, the initial formation is divided into an upper layer of 4 agents and a lower layer of 8 agents. As shown in Fig. \ref{all} (c), two additional agents join in the formation. The ratio of agents in the two layers is reassigned according to the area ratio, followed by PAAS to form a new formation of 14 agents. After traversing the first gap, an upper-layer agent leaves the formation and the system subsequently executes PAAS to reconfigure a new formation. Finally, the formation successively passes through two narrow gaps and quickly restores to the desired formation guided by the DG. In summary, our system handles formation planning with varying numbers in narrow environments.

\subsection{Real-world Experiments}
\label{real_exp}
To validate the effectiveness of our method, we implement the algorithm in a physical aerial swarm, and the architecture is shown in Fig. \ref{real_2}(f).
Each drone has independent modules for control, estimation, perception, and planning. The corresponding hardware includes KAKUTE H7 Mini flight controller for control and MID360 LIDAR for estimation and mapping. The mapping, estimation, and planning are run with an onboard computer JETSON ORIN NX in real-time. 

We conduct three sets of physical experiments, including the joining and leaving of agents in obstacle environments and safe traversal through narrow gaps by affine deformation. The DVS trajectory optimization imposes a velocity constraint that is 2 $m/s$, which is less than the executable speed of drones and ensures reliable trajectory tracking for formation reconfiguration. As shown in Fig. \ref{real_2} (e), an initial triangular formation of 3 drones joined by a new agent at $t=2s$, triggering position partitioning and assignment the scale of the entire formation adaptively increases. The entire formation converges below the formation error threshold $e_{dist}=0.1$ within 3 seconds as shown in Fig. \ref{real_2} (g). Similarly, we demonstrate a 4-agent formation where the bottom-left agent exited at $t=2.3s$, with the remaining three reconfiguring into a new formation. 

In order to highlight the adaptability to narrow environments of our method, we also conducted an experiment on formation crossing the narrow gap. As shown in Fig. \ref{fig1} (a), the initial formation width is set as 3.5$m$, and the narrow gap width is 1.7$m$. This means that the formation needs to deform and reduce its width by 50\%. Our method can cross narrow gaps with a smooth trajectory. At the same time, the initial formation distance between two drones in the direction of formation motion is 1$m$. If the scale of the formation is simply adjusted to navigate through the gap, it can lead to swarm collisions. In contrast, our method can effectively guide formation through the narrow gap via spatiotemporal affine deformation.

\subsection{Point Cloud Broadcast Experiment}
\label{point_exp}
%在本文中，我们系统中的DGE需要使用编队中部分智能体的点云来建立占据栅格地图，满足编队拓扑搜索的大范围地图的需求。为了确定点云广播的频率以及单次广播点云的数量满足实际通讯网络的约束，我们在我们实际的网络条件下测试了两个drone之间通讯不同数量点云的延时，对于相同数量的点云通信我们测试了20次，计算对应的平均延时以及最大最小延时。如图1所示，我们将对应延时绘制成曲线图。
In this paper, the DG requires using point clouds from some agents in formation to construct an occupancy grid map, which meets the need for mapping in DVS topology search. To determine the frequency of point cloud broadcasting and the amount of point clouds per broadcast under real-world communication network constraints, we test the communication delay between two drones for transmitting different sizes of point clouds under our actual network conditions. For point cloud communication of each case, we conduct 20 tests and calculate the corresponding average, maximum, and minimum delays. As shown in Fig. \ref{point}, we provide the corresponding delays in a graph. The frequency of agent trajectory planning in our system is 5Hz. To ensure cooperative stability, the frequency that DG planning requires is 5Hz. We filter out point clouds beyond the maximum local trajectory distance and selectively transmit point clouds within a 120-degree FoV. In this case, the number of perceived point in 0.2$s$ is limited to 2000. As shown in Fig. \ref{point}, the communication delay remains below 0.03$s$ for 2000 points. Accounting for other data communication requirements, we configure the system to communicate 2000 points per message while maintaining a communication frequency below 5Hz.

% \subsection{Discussion}
% \textcolor{blue}{The proposed GRPF sparse graph effectively reduces the computation time for trajectory optimization, while demonstrating remarkable formation maintenance and recovery capabilities in complex environments. However, this work does not investigate formation coordination performance under time-varying topological conditions and the connection rate of the sparse graph relies on empirical selection. 
% Considering the swarm reciprocal avoidance, agents still need to receive the trajectories of all other drones. However, polynomial trajectories can be compactly represented by a small number of parameters, which requires a small communication bandwidth and is less than 10 bps. For larger formation planning, communication complexity can be reduced through swarm grouping \cite{hou}.
% Furthermore, although anomaly rejection mechanisms achieve simultaneous connectivity sparsification and mitigation of anomalous agent impacts, our approach is currently limited to scenarios with a small proportion of abnormal agents. The rejection threshold remains an empirically determined threshold requiring systematic refinement through data-driven threshold adaptation mechanisms. Additionally, we adopt idealized communication assumptions, without considering the communication delay \cite{robustmader}, \cite{chai2021review} and link failure \cite{com1} in practical scenarios.}
\section{Conclusion and future work}
\label{sec:conclusion}
This paper proposes a number adaptive formation planning method based on Deformable Virtual Structures (DVS) to guide formation planning when the number of agents changes in narrow environments. We combine the Lloyd algorithm and the Hungarian algorithm to meet the requirements of partitioning and assignment for irregular formations. Rapid formation recovery and traversal through narrow environments are achieved via continuous deformation guidance. Experiments show that our method supports agents with a 15\% change in quantity to quickly restore formation. Compared with cutting-edge methods, it has advantages in narrow gap crossing success rate, formation recovery speed, and environmental adaptability. 

In future work, we will investigate a fully distributed formation planning system with number and environmental adaptability, and employ lightweight map representations for local environment sharing to reduce communication requirements. We further aim to scale the formation to thousands of agents while maintaining robust formation performance.

% \addtolength{\textheight}{-5.6cm} 
\bibliography{ICRA2025zy}
\end{document}